\documentclass{article}
\usepackage{arxiv}

\usepackage[utf8]{inputenc} 
\usepackage[T1]{fontenc}    
\usepackage{hyperref}       
\usepackage{url}            
\usepackage{booktabs}       
\usepackage{amsfonts}       
\usepackage{nicefrac}       
\usepackage{microtype}      
\usepackage{graphicx}
\usepackage{amsmath}
\usepackage{amssymb}
\usepackage{,wrapfig}

\begin{document}
\title{From Colors to Classes: Emergence of Concepts in Vision Transformers}

\author{Teresa Dorszewski$^1$ \\
    \texttt{tksc@dtu.dk}
    \And
    Lenka Tětková$^1$ \\
    \texttt{lenhy@dtu.dk}
    \And
    Robert Jenssen$^{2,3,4}$\\
    \texttt{robert.jenssen@uit.no}
    \And
    Lars Kai Hansen$^1$\\
    \texttt{lkai@dtu.dk}
    \And
    Kristoffer Knutsen Wickstrøm$^2$\\
    \texttt{kristoffer.k.wickstrom@uit.no} \\
    \\
    1 Department of Applied Mathematics and Computer Science,
    Technical University of Denmark \\
    2 Department of Physics and Technology, UiT The Arctic University of Norway \\
    3 Pioneer Centre for AI, University of Copenhagen, Denmark \\
    4 Norwegian Computing Center, Oslo, Norway \\
    \\
}
\maketitle 
\let\svthefootnote\thefootnote
\newcommand\freefootnote[1]{%
  \let\thefootnote\relax%
  \footnotetext{#1}%
  \let\thefootnote\svthefootnote%
}
\freefootnote{Preprint. Accepted at The 3rd World Conference on eXplainable Artificial Intelligence.}
%
\begin{abstract}
Vision Transformers (ViTs) are increasingly utilized in various computer vision tasks due to their powerful representation capabilities. However, it remains understudied how ViTs process information layer by layer. Numerous studies have shown that convolutional neural networks (CNNs) extract features of increasing complexity throughout their layers, which is crucial for tasks like domain adaptation and transfer learning. ViTs, lacking the same inductive biases as CNNs, can potentially learn global dependencies from the first layers due to their attention mechanisms. 
Given the increasing importance of ViTs in computer vision, there is a need to improve the layer-wise understanding of ViTs. In this work, we present a novel, layer-wise analysis of concepts encoded in state-of-the-art ViTs using neuron labeling. Our findings reveal that ViTs encode concepts with increasing complexity throughout the network. Early layers primarily encode basic features such as colors and textures, while later layers represent more specific classes, including objects and animals. As the complexity of encoded concepts increases, the number of concepts represented in each layer also rises, reflecting a more diverse and specific set of features. Additionally, different pretraining strategies influence the quantity and category of encoded concepts, with finetuning to specific downstream tasks generally reducing the number of encoded concepts and shifting the concepts to more relevant categories.
\end{abstract}
\section{Introduction}

Vision Transformers (ViTs)~\cite{dosovitskiy2020vit} are becoming increasingly important in the field of computer vision~\cite{han2022survey}, but the understanding of their feature extraction process is still in its infancy. This understanding is crucial, as the myriad of learned features form the basis of the network's decision process and can reveal biases~\cite{geirhos2020shortcut} and shortcuts~\cite{ilyas2019adversarial}, providing a deeper understanding of the learning process as a whole~\cite{olah2018building}. One path towards improving this understanding is to investigate where semantic concepts are learned throughout the network. Several such studies exist for convolutional neural networks (CNNs) (e.g.~\cite{alain2016understanding,bau2017netdissect,zeiler2014visualizing}), but very few studies exist for ViTs (e.g.~\cite{vielhaben2024beyond}) or are mainly focused on the output layer and do not analyze what concepts are learned in intermediate layers (e.g.~\cite{bau2017netdissect,clipdissect}). In CNNs, a hierarchical way of processing information is given by the architecture and the constrained receptive fields, which leads to early layers being focused on colors, patterns, and edges while only later layers are able to process complex information like objects and scenes~\cite{bau2017netdissect}. In ViTs, however, this hierarchical information processing is not given by the architectural constraints, as the self-attention layers act globally and can attend to the full image from layer one~\cite{dosovitskiy2020vit}. ViTs could therefore in principle learn complex features already in early layers. While it has been shown that ViTs process information differently and develop more uniform representations compared to CNNs~\cite{raghu2021vision}, it has not been investigated where ViTs start processing complex information and if they focus on colors and patterns in earlier layers just as CNNs do.
Therefore, there is a need for a deeper analysis to understand the layer-wise learning process of ViTs and we propose using concept-based explainability to gain these insights.

In this work, we propose a comprehensive analysis of the layer-wise learning process in ViTs using neuron labeling \cite{clipdissect}. We aim to identify and understand the semantic concepts learned at different layers of the network, providing insights into the hierarchical feature extraction process. We use the CLIP-dissect method introduced by Oikarinen and Weng~\cite{clipdissect}, which provides a fast and reliable method for neuron labeling, due to its flexibility, computational speed, and high quality labels (e.g. shown in a recent quantitative study \cite{kopf2024cosy}).

\begin{figure}[t]
    \centering
    \includegraphics[width=0.8\linewidth]{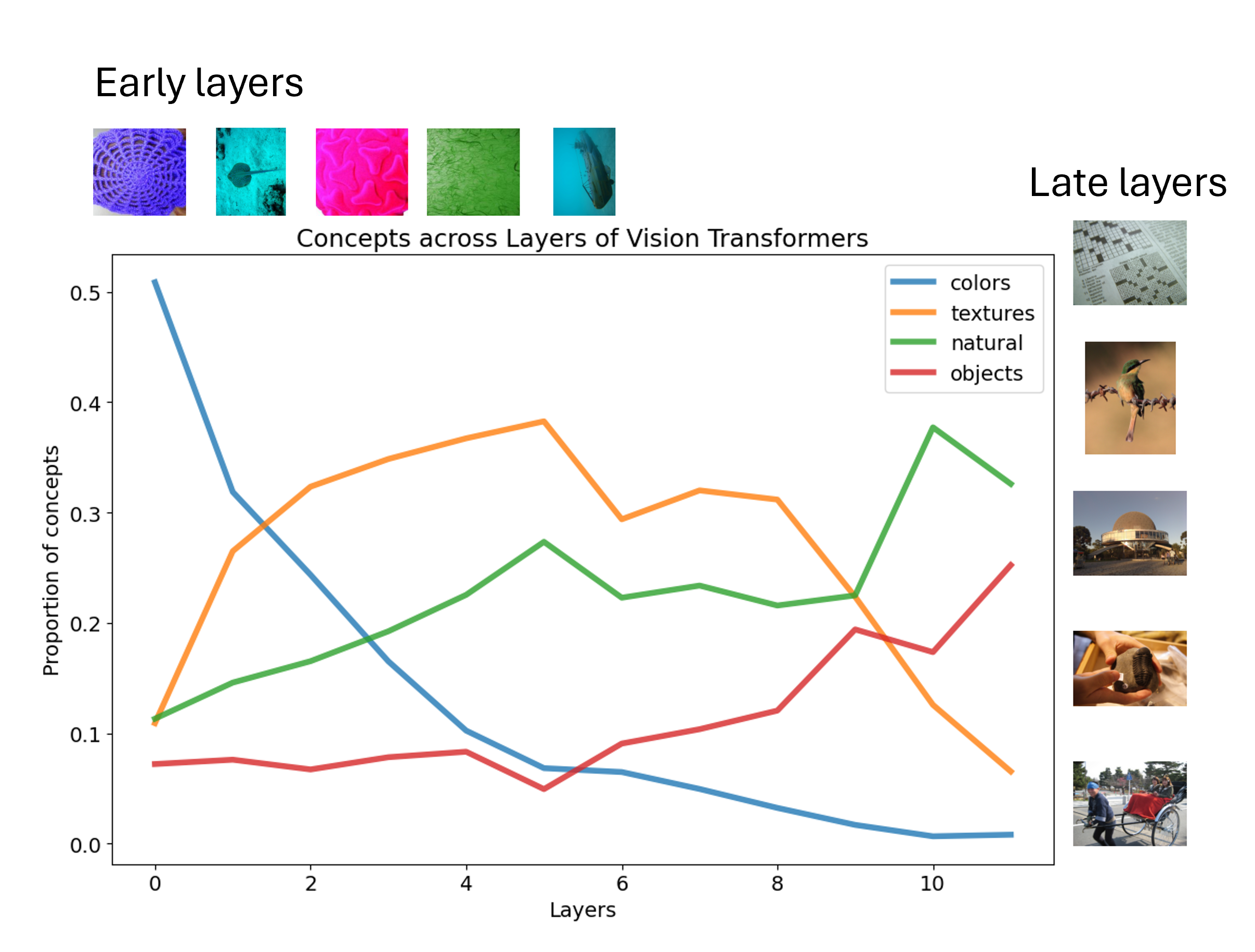}
    \caption{We analyze how different concepts develop across the layers of vision transformers. Early layers tend to process simpler concepts and images while late layers focus on more complex and diverse concepts.}
    \label{fig:1}
\end{figure}

Most of these methods showcase neuron labeling in CNNs, such as ResNets~\cite{he2016resnet}, and have only limited or no experiments on transformer-based models, which are becoming increasingly popular in the computer vision domain \cite{he2022mae,oquab2023dinov2}. The previous works also lack a thorough quantitative analysis of how the concepts develop across layers, and where what kind of information is processed. In this paper, we aim to close this gap by providing a deep analysis of encoded concepts through neuron labeling in ViTs. We find that, despite not being locally constrained, ViTs tend to learn simpler concepts like colors and patterns early on and more complex concepts like objects and natural elements in late layers. \autoref{fig:1} shows how different concept categories develop across layers in ViTs.

Our study highlights the hierarchical feature extraction in ViTs and provides the first comprehensive layer-wise concept analysis of ViTs. Our main findings include: 
\begin{itemize}
    \item ViTs encode more universal concepts, like colors and textures, in early layers and more specialized concepts, like objects and natural elements, and a greater amount of different concepts in later layers.
    \item Neurons in early layers mostly activate on \textit{simpler} images while neurons in late layers react on more \textit{complex} images.
    \item Finetuning to specific downstream tasks reduces the number of encoded concepts and can lead to a concept shift to task-relevant concepts while forgetting irrelevant concepts.
\end{itemize}

\section{Related Work}

Here, we present an overview of investigations from prior works that align the most closely to our work.

\paragraph{Layer-wise concept analysis in CNNs} Numerous studies have investigated the complexity of concepts learned in different layers of the network. An early work by Yosinski et al.~\cite{Yosiniki2014} demonstrated empirically that earlier layers learn to extract more general and low-level features while later layers extract more complex and task-specific features. A later work by Bau et al.~\cite{bau2017netdissect} modeled concepts explicitly through the Network Dissection framework and found that earlier layers extract concepts related to color and textures while later layers extract concepts related to objects. A recent study by Fel et al. \cite{fel2024complexity} used information-theoretic estimators to quantify the complexity of concepts learned throughout the layers of CNNs, which also found that CNNs encode \textit{simple} concepts in early layers and build more \textit{complex} concepts in late layers, with \textit{simple} concepts flowing through residual connection to stay available throughout the network. 

\paragraph{Layer-wise concept analysis in ViTs} To the best of our knowledge, very little work has been conducted in layer-wise concept analysis in ViTs. Recently, Vielhaben et al.~\cite{vielhaben2024beyond} looked into concept-based alignment in ViTs and revealed that early and middle layers encode fewer concepts, that are not well aligned with WordNet categories, while the last layer encodes more concepts than can be semantically organized by WordNet classes (i.e. objects and organisms). Visual inspection revealed that early layers mainly encode colors and structures, however, no quantitative analysis was done. Another notable study by Raghu et al.~\cite{raghu2021vision} investigated the representations of ViTs in comparison with CNNs, which found striking differences in terms of how the similarity of representations are distributed between layers. However, they only looked at the similarity of representations and not into what concepts are encoded in each layer. Two other studies found ViTs to be more consistent with human decision- and error-making and exhibiting less of a texture bias than CNNs, suggesting that ViTs rely on higher-level and more general features and concepts~\cite{naseer2021intriguing,tuli2021convolutional}.

\paragraph{Neuron labeling in computer vision} The analysis in this work is based on the CLIP-Dissect method, due to its high flexibility, high computational speed, and good performance in recent quantitative studies~\cite{kopf2024cosy}. However, several neuron labeling techniques have been proposed recently, and could provide additional perspectives in future studies. The Network Dissection framework ~\cite{bau2017netdissect} was one of the earliest general framework for labeling neurons, The MILAN technique allowed neuron labeling beyond predefined labels and to an open-vocabulary setting~\cite{hernandez2021milan}. INVERT took a compositional concept approach, which eliminated the need for segmentation masks that prior methods relied on~\cite{bykov2024invert}. A recent study did a quantitative evaluation of the quality of the labels \cite{kopf2024cosy}, which found that CLIP-Dissect generally has strong performance.

\section{Theoretical Framework for Neuron Labeling}
\label{sec:theory}
In this section, the CLIP-Dissect method will be described as introduced by Oikarinen and Weng~\cite{clipdissect}. For all experiments, we use the original implementation\footnote{\url{https://github.com/Trustworthy-ML-Lab/CLIP-dissect}}, with only minor changes to support models from the Hugging Face framework \cite{wolf2020transformers}. This is meant to give an overall understanding of how the method works, for more details please refer to the original publication~\cite{clipdissect}. 

The CLIP-Dissect algorithm requires a neural network to be analyzed, in our case different ViTs, a set of probing images ($D_{probe}$) which can be any dataset without the need for concept labels, and a set of concepts ($C$), which is provided as a list of words.
The output of CLIP-Dissect is a set of neuron labels that identify the concept associated with each individual neuron. 

The algorithm consists of three main steps: 
\begin{enumerate}
    \item Compute concept-activation matrix $\textbf{P}$ by computing the $M$ text embeddings $\textbf{t}_j$ and $N$ image embeddings $\textbf{i}_i$ of the concepts $c_j$  in the concept set $C$ and the images $\textbf{D}_i \in \mathbb{R}^{C\times H \times W}$ in the probing set $D_{probe}$ using the text and image encoder of the CLIP model~\cite{radford2021clip}. $\textbf{P} \in \mathbb{R}^{N\times M}$ is a matrix where the inner product of the embeddings gives the $(i,j)$-th element: $\textbf{P}_{i,j} = \textbf{i}_i\cdot \textbf{t}_j$
    \item Record activation of neurons for every image $\textbf{D}_i \in D_{probe}$
    \item Determine neuron label of neuron $k$ by finding the most similar concept $c_m$ (based on mutual information) given the set of highest activating images $B_k$ as activation vector $\textbf{q}_k$ . The label is then given by $\arg\max_m \operatorname{sim}(c_m, \textbf{q}_k; \textbf{P})$. The similarity function is the softWPMI, which is defined as: 
    \begin{equation}
        \operatorname{sim}\left(c_m, \textbf{q}_k; \textbf{P}\right) \triangleq \operatorname{soft\_ wpmi}\left(c_m, \textbf{q}_k\right)=\log \mathbb{E}\left[p\left(c_m \mid B_k\right)\right]-\lambda \log p\left(c_m\right), 
    \end{equation}
\end{enumerate}
where $\log \mathbb{E}\left[p\left(c_m \mid B_{k}\right)\right]=\log \left(\prod_{\textbf{D} \in \mathcal{D}_{\text {probe }}}\left[1+p\left(\textbf{D} \in B_k\right)\left(p\left(c_m \mid \textbf{D}\right)-1\right)\right]\right)$.

The algorithm outputs a label and a similarity score for each neuron of the model. Since we can not expect every neuron to have a clear label, we propose to add adaptive thresholding to the similarity scores to analyze only neurons that have an accurate and trustworthy label. The procedure for this is described in the next section.

\section{Layer-wise Concept Analysis in Vision Transformers}
The code for all experiments and results of the neuron labeling are available at \url{https://github.com/teresa-sc/concepts_in_ViTs}. 

\subsection{Analysis of Concepts}
\label{sec:analysis}
The analysis of concepts is based on the CLIP-dissect method~\cite{clipdissect}, which is described in detail in \autoref{sec:theory}. It outputs a description (``concept'') and similarity score for each neuron in analyzed layers. We analyze the final MLP layer in each transformer block of the ViT models (a total of 12 layers with 768 neurons each) and the last convolutional layer in each residual block in the ResNet50 (a total of 5 layers, with 64, 156, 512, 1024, and 2048 neurons respectively). The exact models are described in \autoref{sec:models}.

To analyze only reliably labeled neurons, we propose to threshold the neurons by the similarity score and analyze only the descriptions of the neurons that have a high similarity score, i.e. a reliable label. According to a user study performed by Oikarinen and Weng~\cite{clipdissect}, around 55-80\% of the descriptions are fitting and around 20-30\% of the neurons do not have a simple label, i.e. are uninterpretable. They suggest using an interpretability cutoff of $\tau=0.16$~\cite{clipdissect}. 

However, the similarity score depends on the concepts and probing dataset. Therefore, we decided to introduce a mean thresholding that can directly be applied to new models and changed datasets. We threshold the descriptions of each layer by the mean of the similarity scores in that layer, which results in slightly different thresholds $\tau$ in each layer of each model, but a comparable amount of labeled neurons so the following quantitative and qualitative analysis is comparable across models and datasets. The exact thresholds and number of neurons with an analyzed description are detailed in \autoref{tab:threshold}.

We analyze how many different concepts are encoded in each layer and which categories these concepts belong to. The number of concepts is determined by simply counting the unique descriptions after thresholding. Each concept is part of a category, as described in \autoref{sec:models}, and we analyze the percentage of labeled neurons that belong to which category. Here we do not only count unique concepts but all neurons that have a concept related to each category, so for example if 10 neurons have the label \textit{blue}, it would count as one concept for counting the amount of different concepts but as 10 \textit{color} neurons. This also ensures that the categories are comparable, even if the amount of different concepts in each category varies. 

\begin{table}[t]
    \centering
    \caption{Mean threshold $\tau$ across layers for each analyzed model and the number of neurons with similarity scores above $\tau$ (average for all vision transformer models, exact number for all layers of ResNet).}
    \begin{tabular}{l|c|c}
model & threshold $\tau$ & \# of labeled neurons \\
\hline \hline
ViT & $0.17 \pm 0.01$ & $317 \pm 13$ \\
\hline
DINOv2 & $0.21 \pm 0.04$ & $323 \pm 11$ \\
\hline
CLIP & $0.17 \pm 0.03$ & $309 \pm 13$ \\
\hline
MAE & $0.17 \pm 0.02$ & $331 \pm 17$ \\
\hline
ResNet50 & $0.21 \pm 0.05$ & $32, 92, 209, 423, 893$ \\
\hline
    \end{tabular}
    \label{tab:threshold}
\end{table}

\subsection{Complexity of Concepts}
We measure the perceived image complexity of the 5 top activating images for each neuron. We estimate the complexity of each image using ICNet~\cite{feng2023ic9600}, which was trained to predict the image complexity scores based on a human-annotated dataset of 9600 diverse images. For each image, we take the mean of the complexity map (a mask with per-pixel complexity scores). The resulting complexity score is a number between 0 and 1, with higher values indicating more complexity. We average these scores across all neurons in each layer.

\subsection{Models and Data Selection for Concept Analysis}
\label{sec:models}
We analyze four commonly used pretrained ViT models, namely the classical supervised Vision Transformer (sup-ViT)~\cite{dosovitskiy2020vit}, DINOv2~\cite{oquab2023dinov2}, Masked Autoencoder (MAE)~\cite{he2022mae} and Contrastive Language-Image Pretraining (CLIP) with a ViT backbone~\cite{radford2021clip}. All models were obtained in their pretrained form from Hugging Face \cite{wolf2020transformers}. The models all share the same architecture of 12 transformer blocks with an embedding size of 768 in each block and take images of the size 224x224 as input. They only differ in their pretraining strategy. Sup-ViT is pretrained in a supervised fashion on ImageNet-21k~\cite{dosovitskiy2020vit}. DINOv2 uses self-supervised learning based on a self-distillation method with pseudo labels. The model is pretrained on a dataset of 142 million images without labels, focusing on producing robust visual features~\cite{oquab2023dinov2}. MAE is pretrained using a self-supervised approach where parts of the input image are masked, and the model learns to reconstruct the missing parts, which helps in learning meaningful representations~\cite{he2022mae}. CLIP is pretrained on a variety of image-text pairs from the internet, learning to predict the most relevant text snippet for a given image, enabling zero-shot capabilities~\cite{radford2021clip}. We also run the same analysis on ResNet50~\cite{he2016resnet} trained on ImageNet-1k for a comparison of ViTs with a CNN. 

To investigate the effect of finetuning, we also analyze sup-ViT, DINOv2, CLIP and ResNet50 finetuned to Caltech-UCSD Birds-200-2011 Dataset (CUB)~\cite{Wah2011cub} and a dataset of the MedMNIST collection~\cite{Yang2023MedMNIST}, the bloodMNIST dataset~\cite{acevedo2020bloodmnist}. The CUB dataset consists of 200 different bird species, which are close to animal and bird images in the pretraining datasets. However, to solve this task, a greater focus on details and parts of the birds is necessary, so it is interesting to investigate how the models and concepts adapt to such fine-grained classification tasks. The bloodMNIST dataset, on the other hand, consists of 8 classes of microscopic peripheral blood cell images, which are very different from the images in the pretraining datasets. This analysis can give insights into how the models and their concepts adapt to new domains. 

The trained models are available on \url{https://huggingface.co/teresas}. All models were trained with a batch size of 32, a scheduler and early stopping for a maximum of 10000 steps, the learning rate was optimized using a validation set. The exact learning rates and final test accuracies can be seen in \autoref{tab:model_training}. 

\begin{table}[t]
 \caption{Learning rate (lr) and test accuracy (acc) in \% for all models trained on CUB and bloodMNIST.}
    \centering
    \begin{tabular}{l|cc|cc}
      & \multicolumn{2}{c|}{CUB} & \multicolumn{2}{c}{bloodMNIST} \\ \hline
     model    & lr         & acc         & lr            & acc     \\ \hline \hline
     sup-ViT      &   $1e^{-4}$   &    0.84     &   $1e^{-4}$   &  0.99  \\
     DINOv2   &   $1e^{-5}$   &    0.87     &   $1e^{-5}$  &   0.99 \\
     CLIP     &   $1e^{-5}$   &    0.80     &   $1e^{-5}$   &  0.98   \\
     ResNet50 &    $1e^{-4}$  &    0.76     &   $1e^{-4}$   &  0.98   \\  
    \end{tabular}
    \label{tab:model_training}
\end{table}

We label the 768 output neurons of each transformer's block final MLP layer using CLIP-dissect~\cite{clipdissect}. For this, we use the ImageNet validation set~\cite{imagenet} and Broden~\cite{bau2017netdissect} as probing dataset (total of $50000+63305=113305$ images) and a list of 20k most common words in English as concept set, as it was defined and used in the original method. 

Out of the 20k concepts, only 1450 different words are used as neuron labels after thresholding the labels across all models. We divide these words into semantic categories, namely colors, textures and materials, objects and machines, places and buildings, natural elements and organisms, activities, names, abstract, and unknown. Each word belongs to exactly one category (see \autoref{tab:categories} for the number of words in each category). This categorization was done by Microsoft Copilot~\cite{copilot} and ChatGPT~\cite{openai2023chatgpt} with manual supervision and adjustments. In case of words fitting to multiple categories, we chose the one subjectively most appropriate. The final list of categories is available at \url{https://github.com/teresa-sc/concepts_in_ViTs/tree/main/data}.

\begin{table}[]
\centering
\caption{Frequency of words in our semi-manual categorization.}
\begin{tabular}{l|c|l}
\hline
Category & Number of words & Examples\\ \hline \hline
Colors & 45 & green, orange, turquoise\\ \hline
Textures and materials & 74  & tiles, dotted, woven\\ \hline
Objects and machines & 449 & furniture, violin, telescope \\ \hline
Places and buildings & 270 & library, Scotland, cabin\\ \hline
Natural elements and organisms & 254 & elephant, mushroom, plant \\ \hline
Activities & 154 & golf, fishing, cooking \\ \hline
Abstract & 127 & itinerary, habit, adorable\\ \hline
Names & 43 & fujifilm, firefox, merlin \\ \hline
Unknown & 34 & aaa, scooby, slashdot \\ \hline
\end{tabular}

\label{tab:categories}
\end{table}

\section{Results}
Here, we present the main results of our analysis. First, we present a layer-wise analysis of numerous state-of-the-art ViTs using the procedure outlined in Section 4. Then, we compare the behaviour of ViTs and CNNs, before we investigate the effect of finetuning on the concepts found in different layers.

\subsection{Layer-wise Analysis of Concepts}
We labeled and analyzed the neurons of four different ViTs using CLIP-dissect~\cite{clipdissect} and found many similarities between the models and a few distinct differences. The full list of labeled neurons for each model is available at \url{https://github.com/teresa-sc/concepts_in_ViTs/tree/main/results}. 

\begin{figure}[t]
    \centering
    \includegraphics[width=0.9\linewidth]{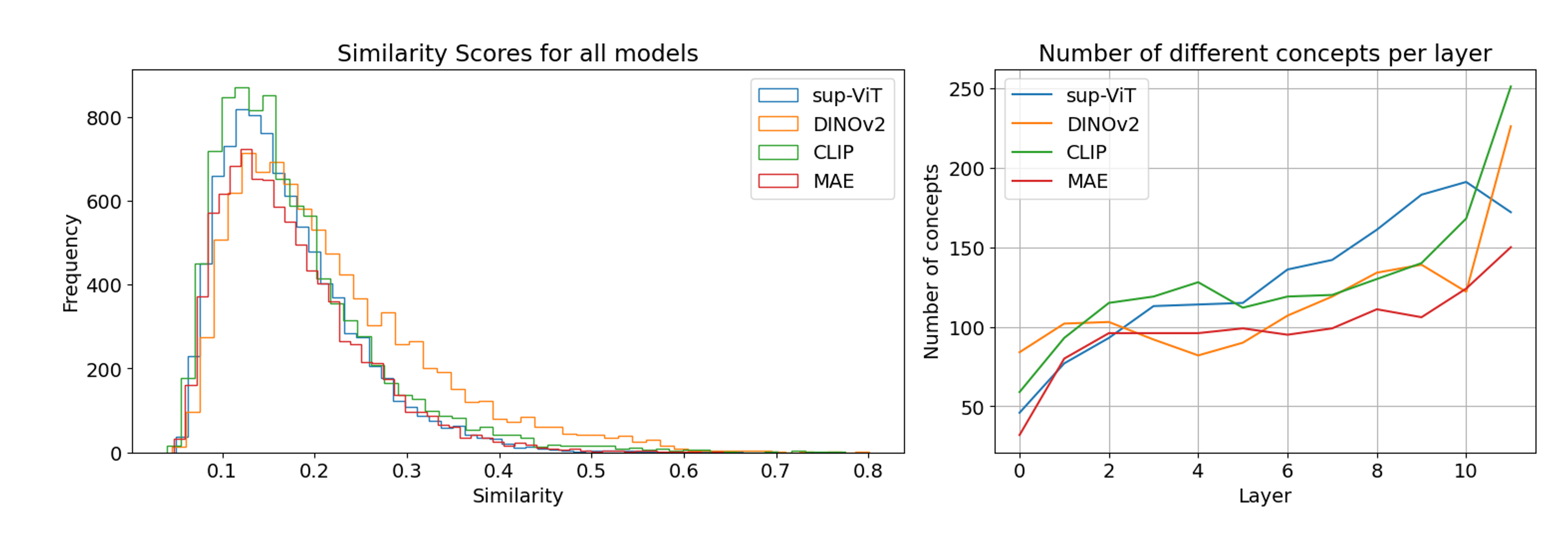}
    \caption{Similarity scores for concepts and number of different concepts encoded in each layer.}
    \label{fig:sim_scores}
\end{figure}

When looking at the overall similarity scores for the description of each neuron all transformer models follow a very similar distribution, as can be seen in \autoref{fig:sim_scores}, with only the DINOv2 model showing overall slightly higher similarity scores. When analyzing the number of different concepts encoded in each layer (after thresholding), a similar trend across models can be observed. At early layers, the models encode fewer concepts (<100) and the number of concepts increases steadily to around double the amount in late layers, the sup-ViT tends to have the most different concepts while the MAE has the least. The DINOv2 follows a slightly different trend with a small decrease of concepts across early layers and a sharp increase in the last layer. 

The smaller amount of concepts is not due to less reliable labels, all layers have a comparable amount of labeled neurons after thresholding (around 310-330), but due to many neurons encoding the same concepts. For example, in the first layer some concepts appear 10-20 times (e.g. ``green'' appears 19 times in sup-ViT layer 1), while in later layers concepts usually only appear once or twice.

\begin{figure}[t]
    \centering
    \includegraphics[width=0.9\linewidth]{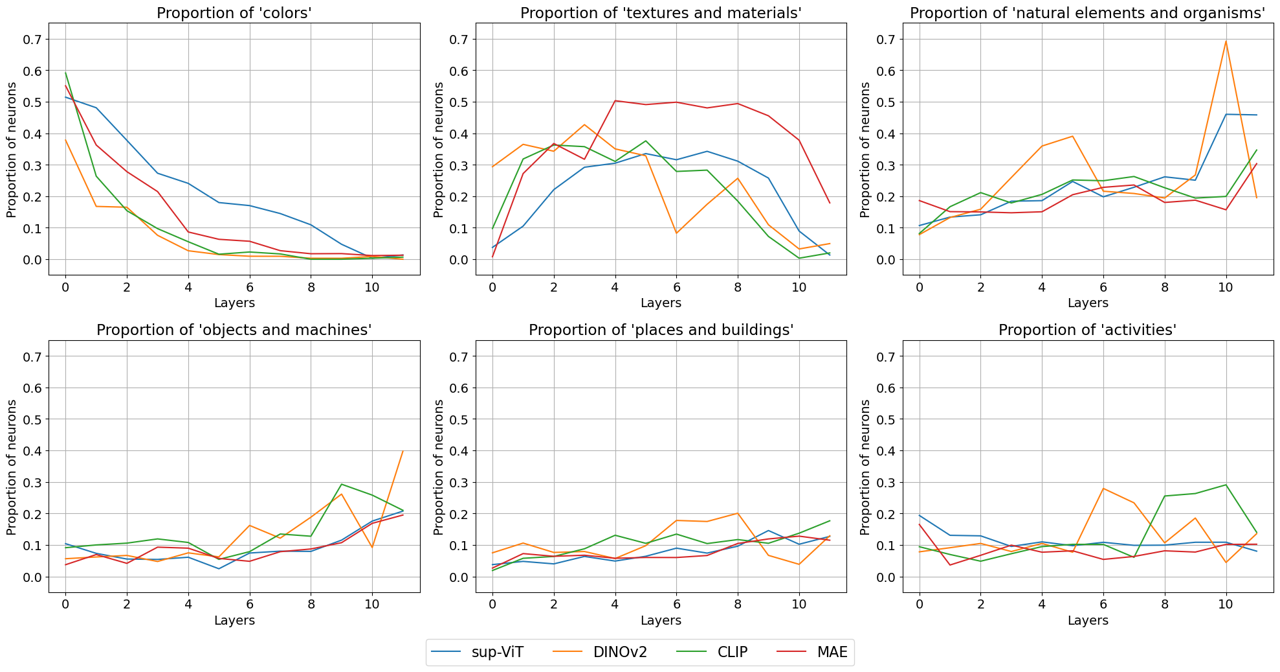}
    \caption{Categories of concepts compared across models. Early layers focus mostly on \textit{colors} while middle layers have a large proportion of neurons assigned to \textit{textures and materials} while \textit{objects} and \textit{natural elements} appear in later layers. The overall trend is similar for all models but small differences can be observed, e.g. CLIP has a higher amount of neurons assigned to activities in late layers.}
    \label{fig:3}
\end{figure}

We break down the concepts into different categories and analyze how they develop across the layers in \autoref{fig:3}, where most concept categories develop similarly across modes. All models start with \textit{colors} as the main concept category present in the first layer, which continually disappears and is almost non-existent in the last layer. \textit{Textures and materials} is also mainly present in early and middle layers and loses importance in late layers but does not disappear completely. Other, more specific categories like \textit{objects and machines} and \textit{natural elements and organisms} start with a small part in early layers and become dominant in late layers. In these categories, we also see some differences between the models. The CLIP model has the highest proportion of \textit{objects and machines} in most layers, but in the last layer, DINOv2 shows a steep increase in this category. All models show a relatively high proportion of \textit{natural elements and organisms}, with DINOv2 and sup-ViT having a very high percentage of neurons belonging to this category in late layers. The DINOv2 model shows a peak in middle layers for the category \textit{places and buildings} and \textit{activities} which other models do not express, they rather show a mostly stable proportion across layers, except for the CLIP model, which has a high proportion ($\sim$25\%) of neurons dedicated to \textit{activities} in late layers. This could be explained by the training strategy of the CLIP model, which is trained by image-caption pairs that might include many activity descriptions. The MAE model seems to have a higher focus on \textit{textures and materials} than other models; it attributes almost half of its neurons in the middle layers to this category. 

Overall, we see a similar pattern of early layers focusing on \textit{simple} concepts like colors, patterns and textures, and late layers specializing to more \textit{complex} concepts like specific objects and natural elements. We excluded the categories \textit{abstract, names} and \textit{unknown} from the detailed analysis and figures as they all have a very small proportion (<5\%) for all layers and models and show no interesting patterns.

\subsection{Complexity of Concepts}

\begin{figure}[t]
    \centering
    \includegraphics[width=1\linewidth]{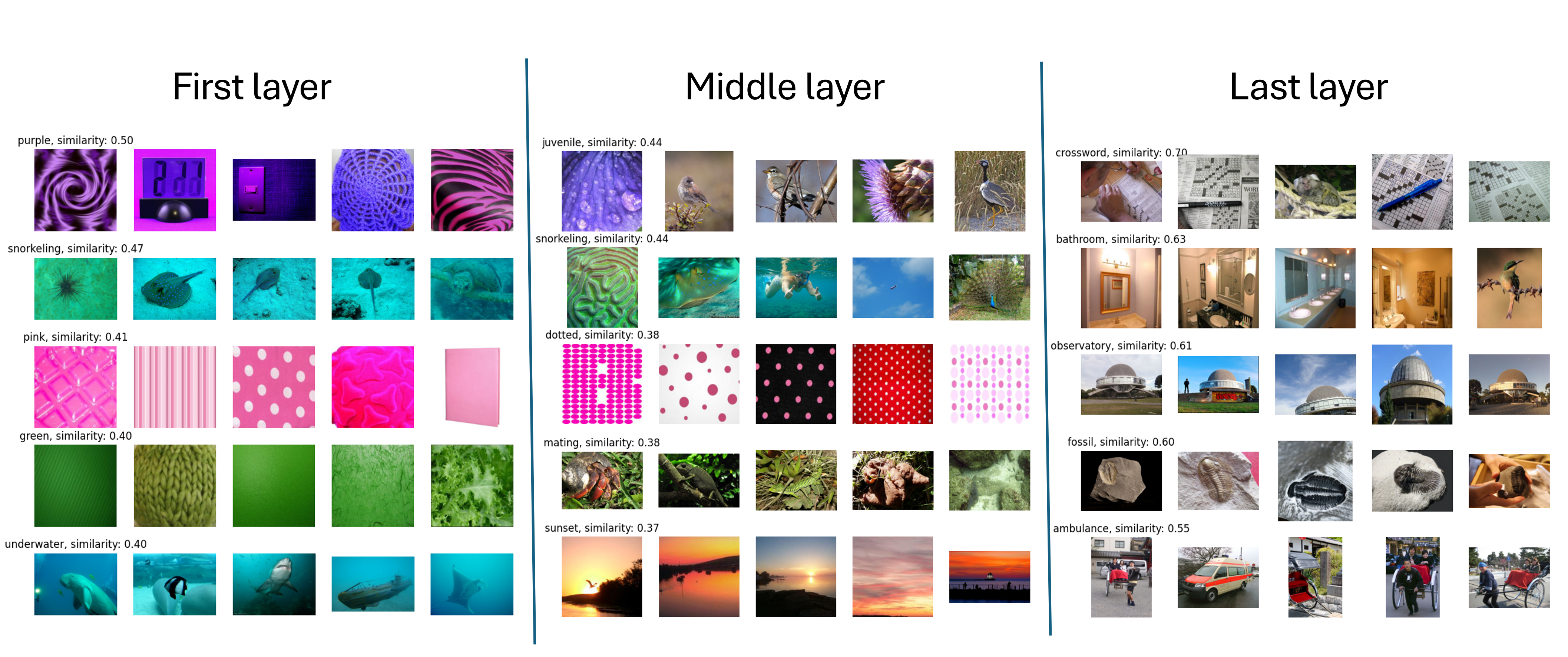}
    \caption{Examples of the highest activating images for five neurons after the first, the sixth and the last transformer block in the sup-ViT model. While this only shows a small subset, the images are representative of most highest activating images, also across models, and show how the models react mainly to simple images in early layers and more complex images in later layers. Similar plots for other models can be found in the Appendix \autoref{fig:app_active_images}.}
    \label{fig:active_images}
\end{figure}

When looking at the top-activating images for each neuron (\autoref{fig:active_images}), it already becomes apparent that early layers mostly activate on \textit{simple} images of mainly colors and patterns while neurons in the last layer mostly activate on more \textit{complex} images of objects and places. Middle layers react on a mix of both with still some \textit{simpler} images mainly driven by color schemes but also on more \textit{complex} images like animals and nature. 

\begin{wrapfigure}{r}{7.5cm}
    \vspace{-15pt}
    \includegraphics[width=1\linewidth]{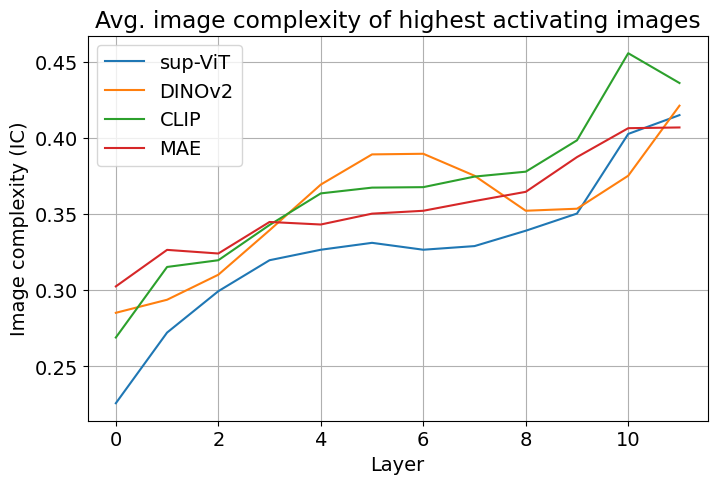}
    \caption{Complexity of the five highest activating images for each neuron averaged across layers measured by ICNet~\cite{feng2023ic9600}. All models show an increase in image complexity across layers.}
    \vspace{-15pt}
    \label{fig:complexity}
\end{wrapfigure}

To not only rely on a subjective interpretation on the complexity of images, we also quantify the image complexity using ICNet~\cite{feng2023ic9600} scoring of the five top-activating images for all neurons and average over each layer (see \autoref{fig:complexity}). We can observe a constant increase in complexity for all models except for DINOv2 which peaks in middle layers and decreases a little before increasing again in the last layer. 

Overall, this analysis in combination with analysis of concept categories shows how ViTs process images in a hierarchal way where early layers focus on basic concepts, like colors and patterns, and late layers specialize on complex and individual concepts. While this has been observed in CNNs as well, where this is enforced by design, ViTs could, in theory, learn complex features from the start, but still seem to build up their information processing in a similar way. We belive our analysis is the first to show this behaviour conclusively for ViTs.

\subsection{Vision Transformers vs. CNNs}
\begin{figure}
    \centering
    \includegraphics[width=\linewidth]{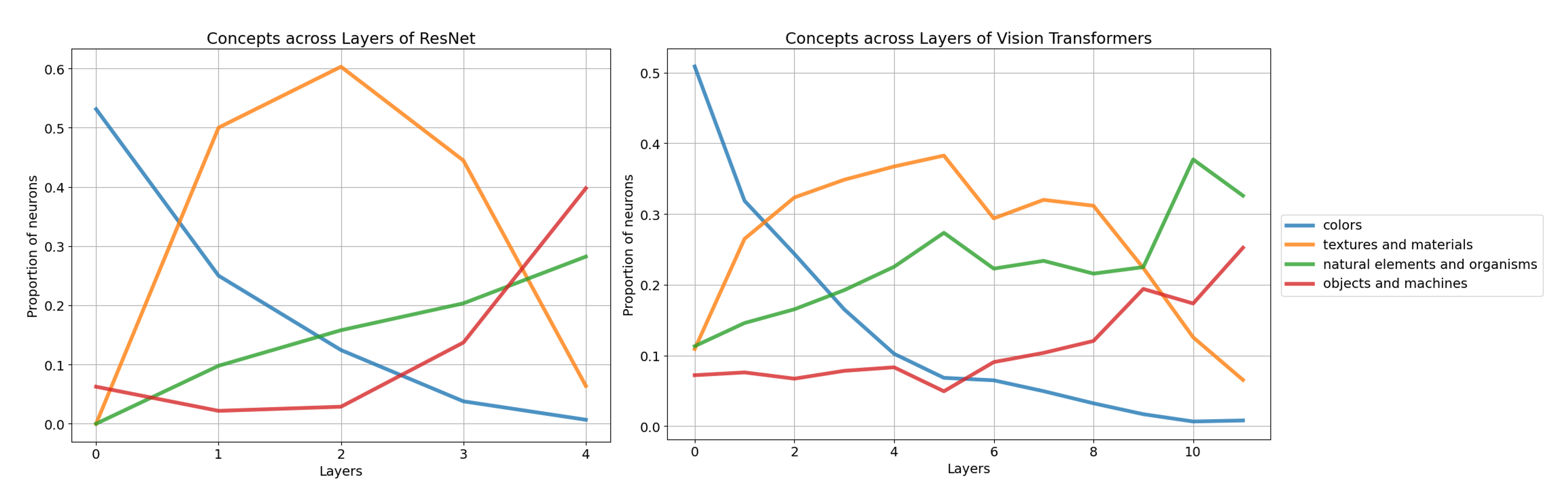}
    \caption{Categories of labeled neurons throughout the models - a comparison of CNN (ResNet50, left) and the average over four ViTs (Vision  Transformer, right) architectures. We see that the general trend is very similar for both architectures.}
    \label{fig:cnn_vs_vit}
\end{figure}

In \autoref{fig:cnn_vs_vit}, we compare categories of the labeled neurons in ResNet50 and the average over all investigated ViTs. The result for ResNet50 is in accordance with previous works for CNNs~\cite{bau2017netdissect,fel2024complexity}. Namely, we see that concepts of \textit{colors} are most important in the beginning of the network, \textit{textures and materials} in the middle, and \textit{objects} and more complicated concepts towards the end, progressing from simple concepts to more complicated ones. The ResNet50 seems to put a higher focus on textures and patterns while the ViTs focus more on objects and natural elements in middle layers. However, the general trend is very similar for both architectures, even though the transformer architecture does not have any inductive bias enforcing this way of processing visual concepts.

\subsection{Change in Concepts after Finetuning}

How do concepts change when a model is finetuned to a specific downstream task? To investigate this, we analyze three ViTs finetuned for two different classification tasks, on the CUB dataset~\cite{Wah2011cub}, which contains 200 bird species, and the bloodMNIST dataset~\cite{acevedo2020bloodmnist}, which comprises of 8 different blood cells. For the classification of birds, it seems obvious that birds and nature elements are important but also colors and patterns could be useful for the classification. For the blood cells, mainly color and shape are relevant. So when finetuning the models to these classification tasks, we might expect more concepts belonging to these relevant categories to appear while other, not so important concepts might be forgotten and disappear. 

\begin{wrapfigure}{r}{7.5cm}
    \vspace{-15pt}
    \includegraphics[width=\linewidth]{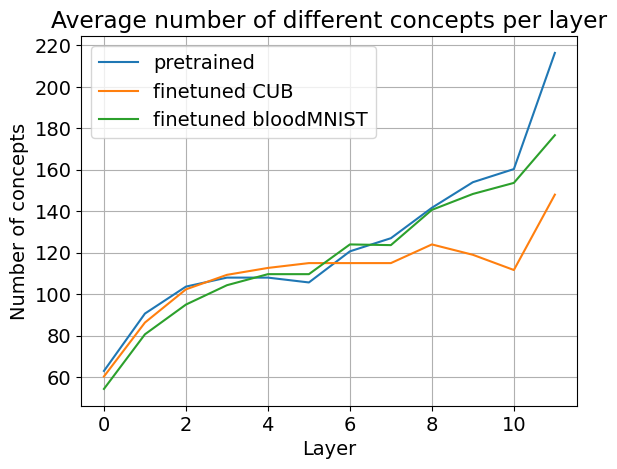}
    \caption{Number of different concepts averaged over three Vision Transformer models in their pretrained and finetuned versions. Finetuning generally decreases the number of different concepts in late layers.}
    \vspace{-15pt}
    \label{fig:num_concepts_ft}
\end{wrapfigure}

\begin{figure}[t]
    \centering
    \includegraphics[width=1\linewidth]{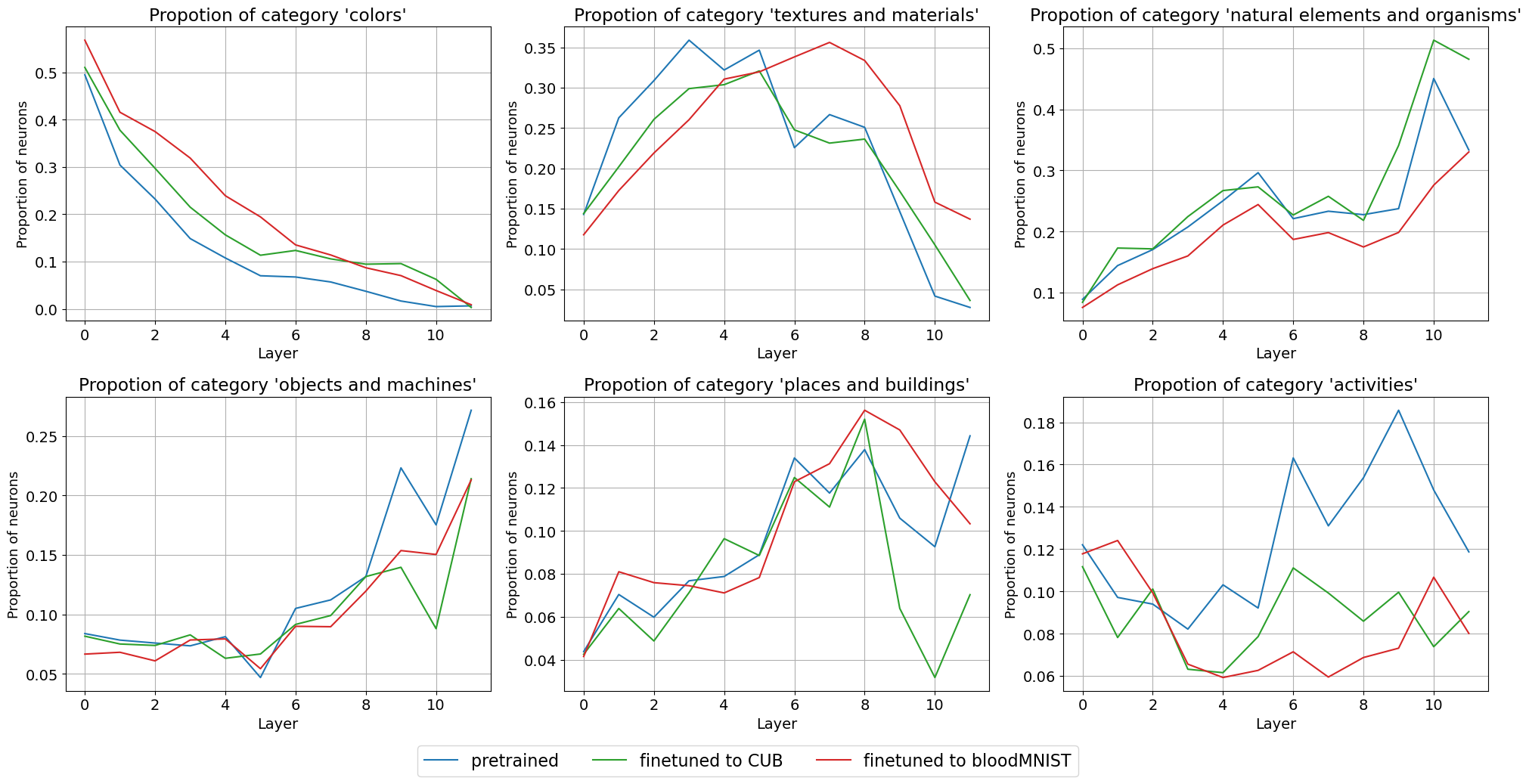}
    \caption{Change in concepts after finetuning averaged over three different Vision Transformer models. The models are finetuned to the CUB dataset and the bloodMNIST dataset and adapt their concepts by putting more focus on relevant categories and ``forgetting'' irrelevant concepts.}
    \label{fig:4_2}
\end{figure}

Indeed, we observe the expected changes. The number of encoded concepts decreases (see \autoref{fig:num_concepts_ft}), and the concepts shift to more relevant categories, as seen in \autoref{fig:4_2}. For models finetuned to bloodMNIST, the category \textit{colors} and \textit{textures and patterns} increase, while \textit{natural elements and organisms}, \textit{objects and machines}, and \textit{activities} decrease. For models finetuned for the CUB dataset, we can also see an increase in the \textit{colors} concepts, but surprisingly only a small increase in \textit{natural elements and organisms}, where any \textit{bird} and \textit{nature} concepts would fall under. The category \textit{textures and materials} seems relatively unchanged while all other categories show a decrease in proportion, as they are not as relevant for the task. Interestingly, when looking at the individual models (see Appendix \autoref{fig:app_ft_models}), we observe that the CLIP model has a very high increase in the \textit{natural elements and organisms} category in late layers while DINOv2 and sup-ViT are almost unchanged. This might be due to these models already being relatively good at distinguishing animals and birds while the CLIP model needs to adapt its weights more to solve this task.  

\section{Discussion}
In this study, we conducted a comprehensive layer-wise analysis of concepts encoded in ViTs using neuron labeling techniques. Our findings provide new insights into the hierarchical feature extraction process of ViTs, revealing how concepts evolve from basic features in the early layers to more complex and specific classes in the later layers, even with the global attention mechanisms that theoretically allow ViTs to learn complex concepts and relations in early layers. We discuss some implications of our findings and highlight future research directions, and also address some limitations of our approach.

\paragraph{On the emergence of distributed representations in ViTs} 
CNNs are designed to learn simpler features and progressively compose them into more complex features, primarily due to the limited receptive field in their early layers. This hierarchical feature extraction process is well-documented, with early layers capturing basic visual elements such as edges and textures and deeper layers encoding more complex structures and objects \cite{bau2017netdissect,fel2024complexity}. In contrast, ViTs can model long-range dependencies from the very first layer, thanks to their self-attention mechanism, which allows them to capture global context across the entire image. Despite this capability, we find that ViTs still tend to learn basic concepts in the initial layers, similar to CNNs. This phenomenon suggests that even with the ability to process global information, the hierarchical nature of feature extraction remains beneficial, and nonlinearities that come with layer-depth might be needed to encode complex concepts. 

We observe that different ViTs learn similar concepts, especially in early layers, where the models learn more general concepts. In later layers, the concepts become more specialized and diverge in some models based on their pretraining strategy. This convergent training on universal concepts has also been observed in CNNs \cite{li2015convergent}.

\paragraph{On the effects of finetuning}
After finetuning we observe a shift in concepts as the model adapts to its new task. This shift to more relevant concepts also results in some concepts being forgotten, which could be linked to the challenge of catastrophic forgetting~\cite{forgetting}. We observe some differences between the models; while DINOv2 and sup-ViT only change marginally when finetuned to CUB, CLIP shows a large shift towards \textit{natural organisms and elements} in the last layers, but still performs worse on the task than the other models. This can be explained by the different pretraining strategies and training datasets and highlights how choosing a model already prepared for the downstreaming task can be beneficial. When finetuning to a more out-of-distribution dataset, e.g. medical images, we observe a larger shift in concepts across all models, which shows how the models can adapt well to new image types, however they most likely lose a greater part of their generalization capabilities during this adaptation. Future work should look deeper into this process and build a greater understanding of catastrophic forgetting using concept-based analysis. 

\paragraph{On the differences of DINOv2} 
While all ViTs we investigated show very similar trends, DINOv2 stands out multiple times with slightly different observations. The neuron descriptions have higher similarity scores than the descriptions for other models, which can be interpreted as the highest activating images having a clearer common description and the neurons potentially being more mono-semantic or specialized. The DINOv2 model also has a higher proportion of neurons reacting to \textit{natural elements and organisms} than other models and a higher complexity of concepts in middle layers.
During finetuning, the concept categories change the least for DINOv2, especially for the CUB dataset, where we see almost no change at all, despite the model performing the best on both tasks compared to the other models. This points to the model already being well-prepared for the downstream tasks and able to generalize to these new datasets without much need for adjusting its weights and encoded concepts. DINOv2 has been shown to outperform many other models and generalize quite well~\cite{oquab2023dinov2}, however, the internal structures that make this possible remain unstudied. A deeper look into how the model processes information and forms concepts can help understand its capabilities, pave the way to even more powerful models, and is a promising direction for future research.

\subsection{Limitations}
\paragraph{Thresholding procedure} 
We chose to employ a mean-thresholding separately on each layer, which leads to a comparable amount of labeled neurons and a higher chance of all investigated neurons having reliable and trustworthy labels, however it also suppresses the chance that some layers and models have more interpretable neurons than others. While for sup-ViT, MAE and CLIP the threshold was $\sim0.17$, which is close to the interpretability cut-off (0.16) determined by a user study \cite{clipdissect}, the mean similarity score for DINOv2 was higher at $\sim0.21$. If we employed a fixed threshold, DINOv2 would stand out with a higher number of concepts, however, the category and finetuning analysis remain unchanged even with a fixed threshold. 

\paragraph{Categorization}
We categorized the concepts with a mix of large language models and manual work, the categories were inspired by previous works and computer vision tasks, however, in the end, subjectively chosen. While we did quality check the categorization, we cannot guarantee there are no mistakes, or others would categorize some concept slightly differently. Therefore, the analysis could look slightly different based on a different categorization. But we are confident, that even with a different categorization or concept set, the main conclusions would not change and the insights gained in our study hold regardless. Future work could include drafting an even more reliable concept set and categorization that would benefit concept-based analysis and explainability in general by providing comparable concept sets across many studies. 

\paragraph{Definition of concepts} 
Many other studies in concept-based explainability define concepts differently then we assume here. Often, concepts are seen as direction in latent space \cite{kim2018interpretability}, specific subspaces \cite{vielhaben2023mcd} or overcomplete dictionaries with which features can be decomposed into concepts \cite{fel2023craft}. We view concepts as concrete labels for neurons, which requires at least some degree of monosematicity of neurons. This has been shown to not be the case for many neurons and models typically encode many more concepts than they have neurons \cite{olah2020zoom}. However, in our analysis, we observe that many neurons do react to a specific type of image, which can be described by common features, i.e. concepts. With our thresholding, we ensure that neurons that are non-interpretable on their own are not included in our analysis. Therefore, we cannot guarantee that we cover all encoded concepts but we can be sure that the concepts we analyze are actually learned by the models and our findings are representative of what neurons react on in each layer and model. Potential future work could include layer-wise analysis using other methods of concept discovery, but many methods are not as easily applied in an automatic and fast way or would need to be adapted to even work on intermediate representations. 

\section{Conclusion}
In this study, we conducted a comprehensive layer-wise analysis of concepts encoded in ViTs using neuron labeling techniques. Our findings provide new insights into the hierarchical feature extraction process of ViTs, revealing how concepts evolve from basic features in the early layers to more complex and specific classes in the later layers. Despite the global attention mechanisms that theoretically allow ViTs to learn complex concepts and relations in early layers, our results show that ViTs still tend to learn basic concepts initially, similar to CNNs. This hierarchical nature of feature extraction in ViTs, even without architectural constraints, suggests that such a hierarchical processing remains beneficial for visual processing and several layers seem to be needed to encode complex concepts.

We observed that different pretraining strategies can influence the category of encoded concepts, and finetuning to specific downstream tasks generally reduces the number of encoded concepts and shifts the concepts to more relevant categories. Especially finetuning to out-of-distribution tasks can cause large concept shifts. 

Overall, our study enhances the understanding of the layer-wise learning process in ViTs and highlights the critical need for future research to focus on concept-based explainability in these models. Future research could further explore the internal structures that enable models like DINOv2 to generalize well and investigate other methods of concept discovery to enhance our understanding of neural representations in ViTs.

\subsection*{Acknowledgments} This work was supported by the Pioneer Centre for AI, DNRF grant number P1, by the Novo Nordisk Foundation grant NNF22OC0076907 ”Cognitive spaces - Next generation explainability”, and by Visual Intelligence, a Centre for Research-based Innovation funded by the Research Council of Norway, grants 309439 and 303514.

%
%
%
\bibliographystyle{splncs04}
\bibliography{refs}

\newpage
\appendix
\section*{Appendix}

\begin{figure}[b]
    \centering
    \includegraphics[width=\linewidth]{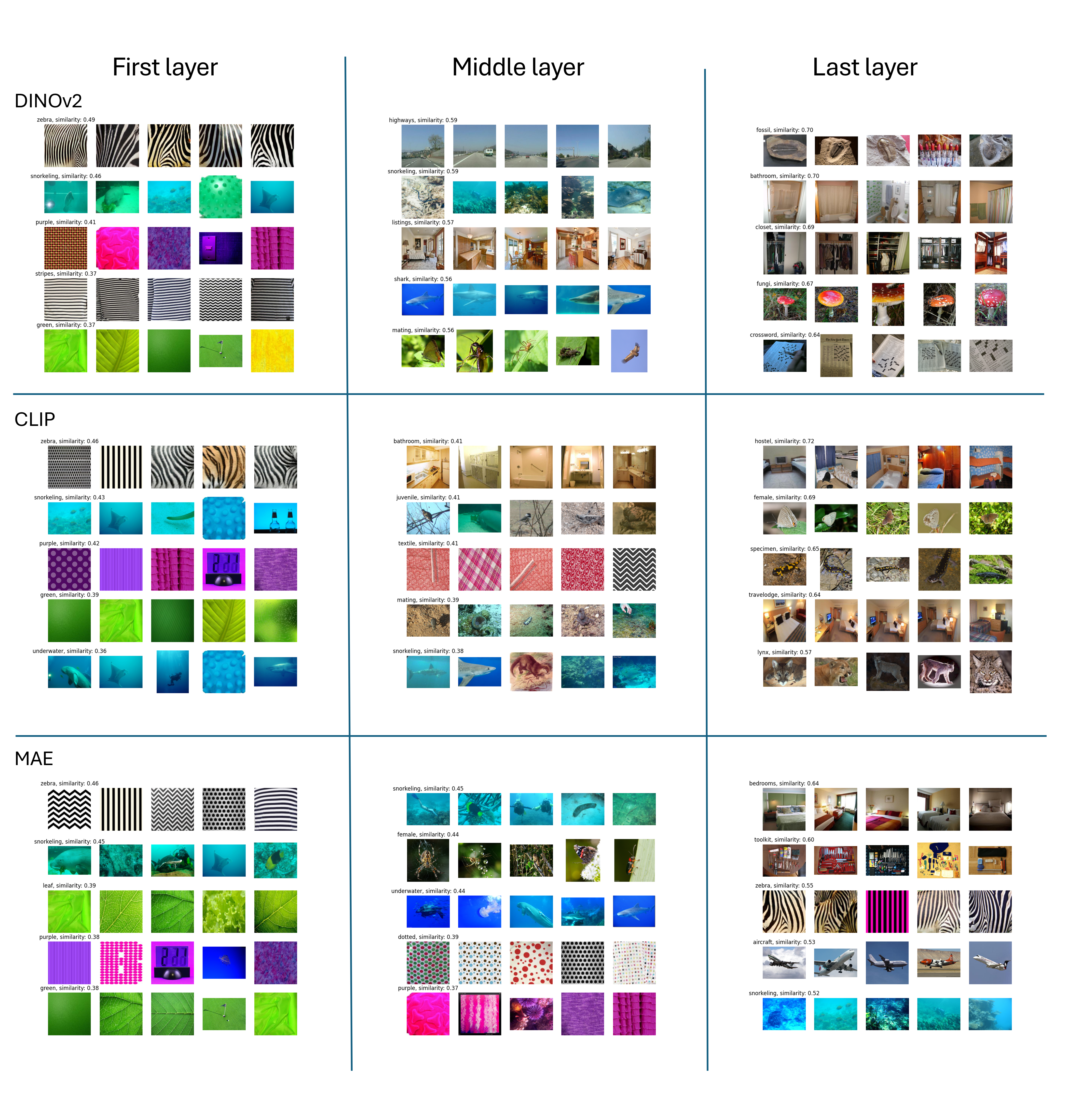}
    \caption{Examples of the highest activating images for DINOv2, CLIP and MAE after layer 0, 6 and 11. Examples for sup-ViT can be seen in \autoref{fig:active_images}.}
    \label{fig:app_active_images}
\end{figure}

In \autoref{fig:app_active_images}, we show some examples of the highest activating images for five neurons after layers 0, 6, and 11. Some differences between models become apparent, e.g. DINOv2 already has more complex images in the middle layers while MAE still has some \textit{simple} textures in the last layer. 

\begin{figure}
    \centering
    \includegraphics[width=\linewidth]{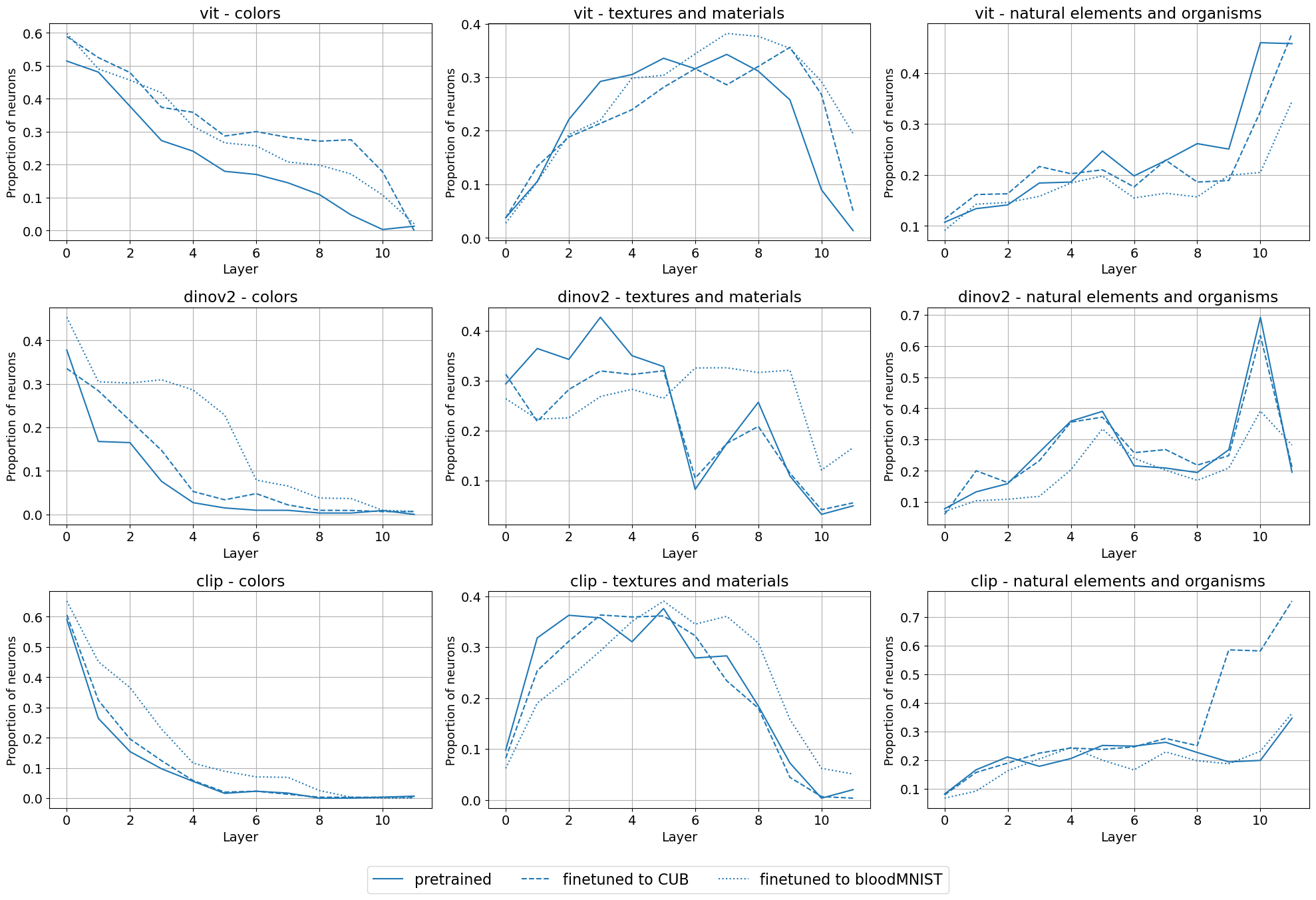}
    \caption{Change during finetuning to the CUB dataset and the bloodMNIST dataset for each model.}
    \label{fig:app_ft_models}
\end{figure}

In \autoref{fig:app_ft_models}, the change in every model during finetuning for the two downstream tasks can be seen. When finetuning to CUB, concepts in DINOv2 change only marginally while CLIP changes a lot for \textit{natural elements and organisms}. For finetuning to bloodMNIST, the overall change is larger, especially DINOv2 changes a lot towards colors and textures. 

\end{document}